\documentclass[10pt,twocolumn,letterpaper]{article}

\usepackage{iccv}
\usepackage{times}
\usepackage{epsfig}
\usepackage{graphicx}
\usepackage{amsmath}
\usepackage{amssymb}
\usepackage{array}
\newcommand{\PreserveBackslash}[1]{\let\temp=\\#1\let\\=\temp}
\newcolumntype{C}[1]{>{\PreserveBackslash\centering}p{#1}}
\newcolumntype{R}[1]{>{\PreserveBackslash\raggedleft}p{#1}}
\newcolumntype{L}[1]{>{\PreserveBackslash\raggedright}p{#1}}

\usepackage[pagebackref=true,breaklinks=true,letterpaper=true,colorlinks,bookmarks=false]{hyperref}

\iccvfinalcopy 

\def\iccvPaperID{3101} 

\ificcvfinal\fi
\begin{document}

\title{Image2song: Song Retrieval via Bridging Image Content and Lyric Words}

\author{Xuelong Li$^*$, ~~~Di Hu$^\dagger$, ~~~Xiaoqiang Lu$^*$\\
$^*$Xi'an Institute of Optics and Precision Mechanics, \\
Chinese Academy of Sciences, Xi'an 710119, P. R. China\\
$^\dagger$School of Computer Science and Center for OPTical IMagery Analysis and Learning (OPTIMAL),\\
Northwestern Polytechnical University, Xi'an 710072, P. R. China\\
{\tt\small xuelong\_li@opt.ac.cn, hdui831@mail.nwpu.edu.cn, luxiaoqiang@opt.ac.cn}
}

\maketitle
\thispagestyle{empty}

\begin{abstract}
Image is usually taken for expressing some kinds of emotions or purposes, such as love, celebrating Christmas.
There is another better way that combines the image and relevant song to amplify the expression, which has drawn much attention in the social network recently.
Hence, the automatic selection of songs should be expected.
In this paper, we propose to retrieve semantic relevant songs just by an image query, which is named as the image2song problem.
Motivated by the requirements of establishing correlation in semantic/content, we build a semantic-based song retrieval framework, which learns the correlation between image content and lyric words.
This model uses a convolutional neural network to generate rich tags from image regions, a recurrent neural network to model lyric, and then establishes correlation via a multi-layer perceptron.
To reduce the content gap between image and lyric, we propose to make the lyric modeling focus on the main image content via a tag attention.
We collect a dataset from the social-sharing multimodal data to study the proposed problem, which consists of (image, music clip, lyric) triplets.
We demonstrate that our proposed model shows noticeable results in the image2song retrieval task and provides suitable songs.
Besides, the song2image task is also performed.
\end{abstract}

\section{Introduction}
Images are usually taken for the purpose of memorizing, which could contain some specific contents and convey some kinds of emotions.
For example, when celebrating the Christmas day, the captured pictures commonly contain dressed up people and Christmas trees covered with gifts, which is used to remember the happiness time.
However, images appear to exist in only visual modality, which could be weak in expressing the above purpose.
Inspired by visual music and musical vision~\cite{li2008visual}, we consider that the stimulates come from different senses (e.g. vision, auditory, tactile) may share similar performance.
Hence, if the captured image is combined with a relevant song that expresses similar purpose, the expression will be obviously enhanced, which results from the more useful merged information from multimodal data~\cite{stein1993merging, Liem2012WhenMM}.
For example, showing the Christmas image while playing the song \emph{Jingle Bells} is easier to touch viewers than single image.
Hence, this kind of combination has attracted much attention in nowadays, which is simpler than video but richer than photo.
But existing approaches of song selection for a given image are almost based on manual manner. Such methods often cost users a lot of time to decide but could still suffer from the small song library of users and lack of song comprehension. Hence, the technique of automatic image-based song recommendation should be expected, which is named as image2song in this paper.

\begin{figure}[t]
\centering
\includegraphics[width=8cm]{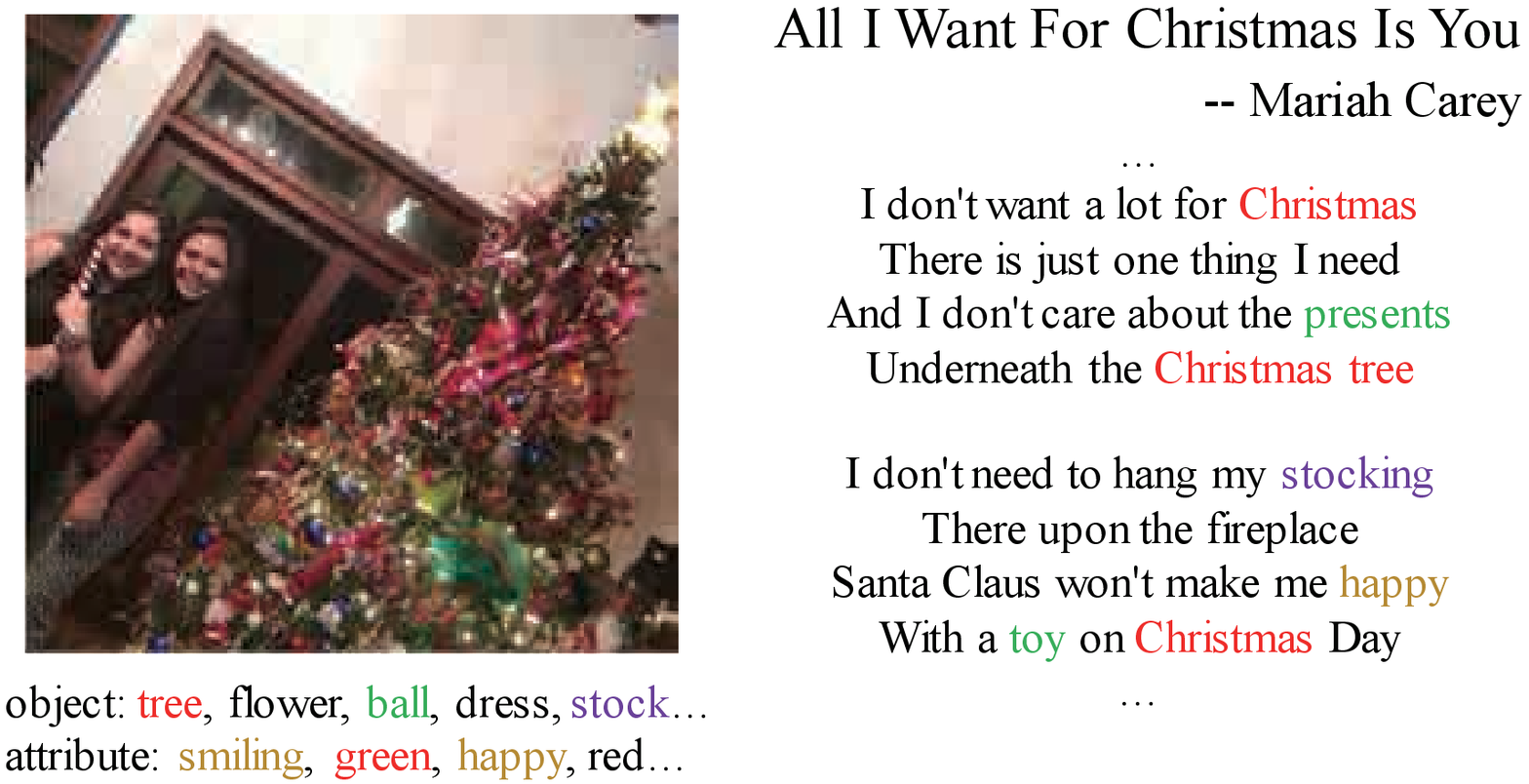}\\
\caption{A Christmas-relevant image and coupled song (lyric). There are several words (in different color) appearing in both image tags and lyric words. Best viewed in color.}\label{illustration}
\end{figure}

\emph{Music Information Retrieval} (MIR) is a traditional research field, which focuses on indexing proper music according to specific criteria.
In this paper, the proposed image2song retrieval task aims to find the semantic-related songs for images, which therefore relies on the specific image content analysis.
Hence, this task belongs to the semantic/tag-based music retrieval category~\cite{schedl2014music}.
Such retrieval tasks utilize multiple textual data sources as the music modality for meeting the semantic requirements, such as music blog in~\emph{SerachSounds}~\cite{celma2006search,sordo2009querybag} and web pages in~\emph{Gedoodle}~\cite{knees2007music}. In this paper, we focus on retrieving songs (not instrumental music) for images, which contain sufficient textual data in lyric. More importantly, these textual data contains multiple common words in image tags (as shown in Fig.~\ref{illustration}), which are considered as the related content across the modalities. Hence, lyric is adopted as the textual data source for retrieval.
However, there still remains another two problems.
First, image and lyric are different, where the former is non-temporal and the latter is temporal-based. More importantly, lyrics are not specifically created for images, which results in the content description gap between them~\cite{schedl2014music}.
Second, there is barely dataset providing corresponding images and songs, which makes it difficult to learn the correlation via a data-driven fashion.

In this paper, to overcome the above problems, our contributions are threefold:
\begin{itemize}
\setlength{\itemsep}{0pt}
\setlength{\parsep}{0pt}
\setlength{\parskip}{0pt}
  \item We leverage the lyric as the textual data modality for semantic-based song retrieval task, which provides an effective way to establish the correlation between image and song in semantic.
  \item We develop a multimodal model based on neural network, which learns the latent correspondence by embedding the image and lyric representations into a common space.
  To reduce the content gap, we introduce a tag attention approach that makes the bidirectional~\emph{Recurrent Neural Network} (RNN) of lyric focus on the main content of image regions, especially for the related lyric words.
  \item We build a dataset that consists of (image, music clip, lyric) triplets, which are collected from the social-sharing multimodal data on the Internet\footnote{The dataset is available at https://dtaoo.github.io/}.
\end{itemize}
Experimental results verify that our model can provide noticeable retrieval results.
In addition, we also perform the song2image task and our model has an improvement over state-of-the-art method in this task.

\section{Related Work}
\noindent \textbf{Image description.}
An explicit and sufficient description of image is necessary for establishing the correlation with lyric in content.
There have been many attempts over the years to provide detailed and high level description of images effectively and efficiently.
Sivic and Zisserman~\cite{sivic2003video} represented the image by integrating the low-level visual features into bag-of-visual-words, which has been widely applied in the scene classification~\cite{li2007and} and object recognition~\cite{sivic2005discovering}.
Recently, in view of the advantages of~\emph{Convolutional Neural Network} (CNN) in producing high-level semantic representation,
many approaches based on it have shown great success in image understanding and description~\cite{russakovsky2015imagenet, Long2015FullyCN}.
However, these methods just focus on describing the image in specific fixed labels, while our work aims at providing a richer description of images, which points out the detailed image contents.
Wu~\emph{et al.}~\cite{wu2015value} proposed an attribute predictor with the same purpose, which viewed the prediction as a multi-label regression task but without focusing on the image contents in specific regions.

\noindent \textbf{Lyric modeling.}
A number of approaches have been proposed to extract the semantic information of lyrics.
Most of these works viewed lyric as a kind of text, therefore~\emph{Bag-of-Words} (BoW)  approach was usually used to describe the lyrics~\cite{chen2006content, hu2009lyric, kim2010music}, which accounted for the frequency of word across the corpus, \emph{e.g.}~\emph{Term Frequency-Inverse Document Frequency} (TF-IDF)~\cite{van2010automatic}.
However, almost all these works aimed at emotion recognition~\cite{kim2010music} or sentiment classification~\cite{xia2008sentiment}.
Recently, Schwarz~\emph{et al.}~\cite{schwarz2016autoIllustration} proposed to embed the lyric words into vector space and extract relevant semantic representation. But, this work just focused on sentences rather than the whole lyric. What is more, Schwarz~\emph{et al.}~\cite{schwarz2016autoIllustration} just performed pooling operation over the sentence words, which ignored to model the contextual information of lyric.
Compared with the previous works, our work focuses on the lyric content analysis for providing sufficient information to learn the correlation with image, which takes consideration of both the semantic and context information of words.

\noindent \textbf{Multimodal learning across image and text.}
Several works have studied the problem of annotating image with text.
Srivastava and Salakhutdinov~\cite{srivastava2012multimodal} proposed to use ~\emph{Multimodal Deep Boltzmann Machine} (MDBM) to jointly model the images and corresponding tags, which could be used to infer the missing text (tags) from image queries, or inverse.
Sohn~\emph{et al.}~\cite{sohn2014improved} and Hu~\emph{et al.}~\cite{hu2016multimodal} extended such framework to explore and enhance the similarity across image and text.
Recently, amounts of works focused on using natural language to annotate image instead of tags, which is usually named as image caption.
These works~\cite{kiros2014multimodal, kiros2014unifying,mao2014deep} commonly utilized deep CNN and RNN to encode the image and corresponding sentences into a common embedding space, respectively.
The aforementioned works aimed to generate relevant tags or sentences to describe the content of images.
In contrast, our work focuses on learning to maximize the correlation between image and lyric, where the lyric is not specially generated for describing the concrete image content but a text containing several related words.

\noindent \textbf{Music visualization.}
Music visualization has an opposite goal of our work, which aims at generating imagery to depict the music characteristic.
Yoshii and Goto~\cite{yoshii2008music} proposed to transform musical pieces into a color and gradation visual effects, \emph{i.e.}, a visual thumbnail image.
Differently, Miyazaki and Matsuda~\cite{miyazaki2010dynamicicon} utilized a moving icon to visualize the music.
However, these methods are weak in conveying semantic information in music, such as content and emotion.
A possible way to deal with such defects is to learn the semantic from lyrics and establish correlation with real images.
Cai~\emph{et al.}~\cite{cai2007automated} manually extracted the salient-words from lyrics, \emph{e.g.} location and non phrases, then took them as the key words to retrieve images. Similar frameworks could be found in~\cite{Wang2007RetrievingWI, Xu2008AutomaticGO}, while Schwarz~\emph{et al.}~\cite{schwarz2016autoIllustration} focused on the sentences of lyrics and organized the retrieved images of each sentence into a video.
In addition, Chu~\emph{et al.}~\cite{chu2016song} attempted to sing the generated description of an image for the first time, but it was too limited for various music style and lack of natural melody.
In contrast to these methods which are based on rough tag description~\cite{cai2007automated, Wang2007RetrievingWI, Xu2008AutomaticGO, schwarz2016autoIllustration} and direct similarity measure between features of different modalities~\cite{schwarz2016autoIllustration}, we propose a multimodal learning framework to jointly learn the correlation between lyric and image in a straightforward way while analyzing the specific content for both of them.

\section{The Shuttersong Dataset}
In order to explore the correlation between image and song (lyric), we collect a dataset that contains amounts of pairwise images and songs from the Shuttersong application.
Shuttersong is a social sharing software, just like Instagram\footnote{www.shuttersong.com, www.instagram.com}. However, the shared content contains not only an image, but also a corresponding song clip selected by users, which is for strengthening the expression purpose. A relevant mood can also be appended by users.
We collect almost the entire updated data from Shuttersong, which consists of 36,646 pairs of images and song clips. Some optional mood and favorite count information are also included.

In this paper, the lyric is viewed as a bridge connecting image and song, but it is not contained in the collected data from Shuttersong.
To acquire the lyrics, we develop a software to automatically search and download them from the Internet based on song title and artist.
However, there exist some abnormal ones in the collected lyrics, such as non-English songs, undetected ones, etc.
Hence, we ask twenty participants to refine the lyrics. Specifically, the participants first exclude the non-English song, then judge whether the lyric matches the song clip. For the incorrect or undetected ones, the participants manually search the lyrics via the Internet. Then, the refined lyrics update the original ones in the dataset. A detailed explanation of the collected data can be found in the supplementary material.

\begin{figure}[h]
\centering
\includegraphics[width=8cm]{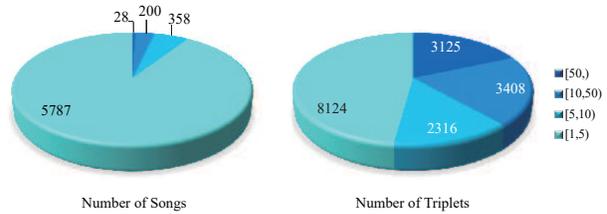}\\
\caption{The statistics of the frequency of song occurrence. For example, there are 5,787 songs appearing less than 5 times, which results in 8,124 triplets.}\label{dataset_pie}
\end{figure}

\begin{figure}[b]
\centering
\includegraphics[width=8cm]{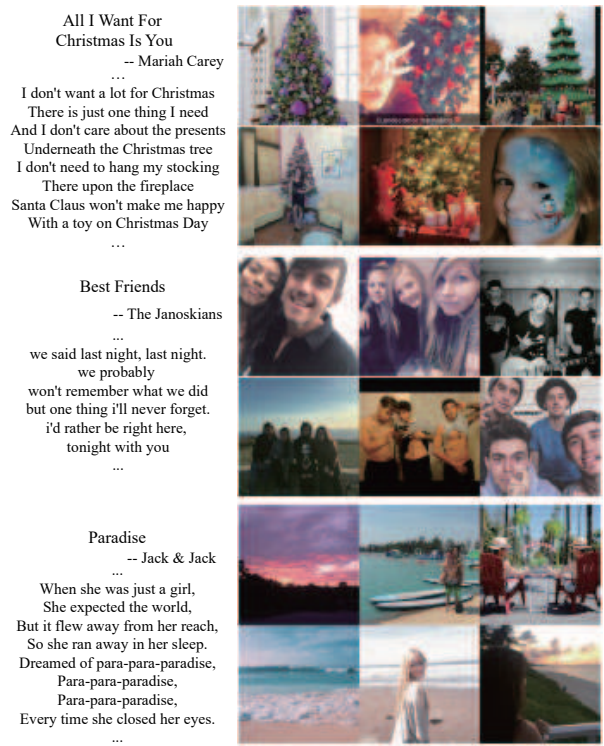}\\
\caption{Examples of songs and corresponding images in the dataset. One song could relate to multiple images. It is easy to find out the images belonging to the same song have similar content that expresses the song to some extent.}\label{dataset_ill}
\end{figure}

\noindent \textbf{Statistics and analysis.}
The full dataset consists of 16,973 available triplets after excluding the abnormal ones, where each triplet contains corresponding music clip\footnote {Due to the relevant legal, it is difficult to obtain the complete audio.}, lyric, and image.
As shown in Table \ref{dataset}, there are totally 6,373 different songs among these triplets, but only 3,158 (18.6\%) ones have available moods.
The favorite counts of the shared image-song pairs created by users vary from 1 to 8,964, which could be used as a reference for estimating the quality of the pairs.

\begin{table}[t]
\begin{center}
\begin{tabular}{c|c|c|c}
  \hline
  Triplet & Song  & Available Mood  & Favorite Count \\ \hline
  16,973 & 6,373 & 3,158 & [1, 8964] \\ \hline
\end{tabular}
\end{center}
\caption{\label{dataset} Aggregated statistics of the Shuttersong dataset.}
\end{table}

We also perform statistical analysis about the frequency of song occurrence, as shown in Fig.~\ref{dataset_pie}. Although there are 6,373 different songs in the dataset, 586 songs that appear at least 5 times take up more than half the triplets, which means each of these songs relates to at least 5 images.
For example, Fig.~\ref{dataset_ill} shows some examples of image-song pairs, where these songs appear more than 5 times in the built dataset.
It is obvious that the images belong to the song \emph{All I Want For Christmas Is You} are Christmas relevant, where trees, ribbons, and lights commonly appear among them.
Meanwhile, these objects or attributes are related to some words of the corresponding lyric to some extent and provide a similar expression with the song.
We can also find the similar situations in the other two groups, as shown in Fig.~\ref{dataset_ill}.
These relevant images of the same song could provide an efficient way to explore the valuable information in lyrics and establish correlation with songs.
Therefore, we conduct our experiments based on these songs with high occurrence frequency.

\section{Our Model}
In this paper, our proposed model aims at learning the correlation between images and lyrics, which can be used for song retrieval by image query, and vice versa.
We first introduce a CNN-based model for fully representing image with amounts of tags, then encode the lyric sequence into a vector representation via a bi-directional LSTM model.
A proposed multimodal model finally embeds the encoded lyric representation into the image tag space to correlate them under tag attention.

\subsection{Image representation}
\begin{figure}[b]
\centering
\includegraphics[width=8cm]{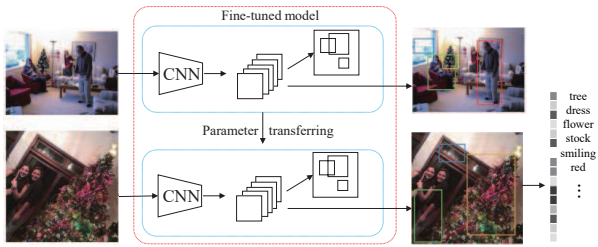}\\
\caption{Overview of the image tag prediction network. The model is developed based on faster-rcnn, which is firstly fine-tuned on the Scene Graph Dataset~\cite{johnson2015image}, then used to generate the tag prediction result for images in the Shuttersong dataset via parameter transferring.}\label{image_frame}
\end{figure}

We aim to generate a rich description of image, which could point out the specific content in certain text, \emph{i.e.}, image tags.
A common strategy is to detect objects and their attributes via extracting proposal regions and feeding them into specific classifier~\cite{kulkarni2013babytalk}.
Hence, we propose and classify the regions of each image via a common~\emph{Region CNN} (RCNN)~\cite{ren2015faster}.
However, different from most of the previous works that just make use of the CNN features in the top K regions \cite{karpathy2015deep, xu2015show}, we use the whole prediction results to provide a richer description.
We adopt the powerful VGG net~\cite{simonyan2014very} as the CNN architecture.
As shown in Fig.~\ref{image_frame}, the CNN is first initialized from ImageNet, then fine-tuned on the Real-World Scene Graphs Dataset~\cite{johnson2015image} that provides more object and attribute classes compared with COCO 2014~\cite{lin2014microsoft} and Pascal VOC~\cite{everingham2010pascal}, where 266 object classes and 249 attribute types\footnote{Different from the settings in~\cite{johnson2015image}, we select the attributes appear at least 30 times to provide a more detailed description.} are employed for fine-tuning, respectively.
The images in the Shuttersong dataset are then fed into the network. The top prediction probabilities of each class constitute the final image representation as a 515-dimensional vector ${v}$.

\subsection{Lyric representation}
Considering the song lyric is a kind of text containing dozens of words, we expect to generate a sufficient and efficient representation to establish inner-modalities correlation with the image representation.
RNN-like architectures have recently shown advantages in encoding sequence while containing enough information for a range of natural language processing tasks, such as language modeling~\cite{sutskever2014sequence} and translation~\cite{graves2012neural}.
In our work, in view of the remarkable ability of LSTM~\cite{hochreiter1997long} in encoding long sequence, we employ it to embed the lyric into a vector representation but with minor modification~\cite{graves2013speech}.

Words of lyric encoded in one-hot representations are first embedded into a continuous vector space, where the nearby points share similar semantic,
\begin{equation}\label{embedding}
{{\rm{x}}_t} = {{\rm{W}}_e}{{\rm{l}}_t},
\end{equation}
where ${\rm{l}}_t$ is the $t$-th word in the song lyric, and the embedding matrix ${{\rm{W}}_e}$ is pre-trained based on the part of Google News dataset (about 100 billion words)~\cite{mikolov2013distributed}.
The weight ${{\rm{W}}_e}$ is kept during the training due to overfitting concerns.
Then the word vectors constitute the lyric matrix representation and are fed into the LSTM network,
\begin{eqnarray}\label{lstm}
{i_t} &=& \sigma \left( {{{\rm{W}}_i}{{\rm{x}}_t}{\rm{ + }}{{\rm{U}}_i}{{\rm{h}}_{t - 1}}{\rm{ + }}{{\rm{b}}_i}} \right)\\
{f_t} &=& \sigma \left( {{{\rm{W}}_f}{{\rm{x}}_t}{\rm{ + }}{{\rm{U}}_f}{{\rm{h}}_{t - 1}}{\rm{ + }}{{\rm{b}}_f}} \right)\\
{{\tilde C}_t} &=& tanh \left( {{{\rm{W}}_c}{{\rm{x}}_t}{\rm{ + }}{{\rm{U}}_c}{{\rm{h}}_{t - 1}}{\rm{ + }}{{\rm{b}}_c}} \right)\\
{C_t} &=& {i_t} * {{\tilde C}_t} + {f_t} * {C_{t - 1}}\\
{o_t} &=& \sigma \left( {{{\rm{W}}_o}{{\rm{x}}_t}{\rm{ + }}{{\rm{U}}_o}{{\rm{h}}_{t - 1}} + {{\rm{b}}_o}} \right)\\
{{\rm{h}}_t} &=& {o_t} * tanh \left( {{C_t}} \right).
\end{eqnarray}
Three gates  (input $i$, forget $f$, and output $o$) and one cell memory $C$ constitute a LSTM cell and work in cooperation to determine whether remembering or not. $\sigma$ is the sigmoid function.
And the network parameters ${{\rm{W}}_*}$, ${{\rm{U}}_*}$, and ${{\rm{b}}_*}$ will be learned during the training procedure.

\begin{figure*}[t]
\centering
\includegraphics[width=17cm]{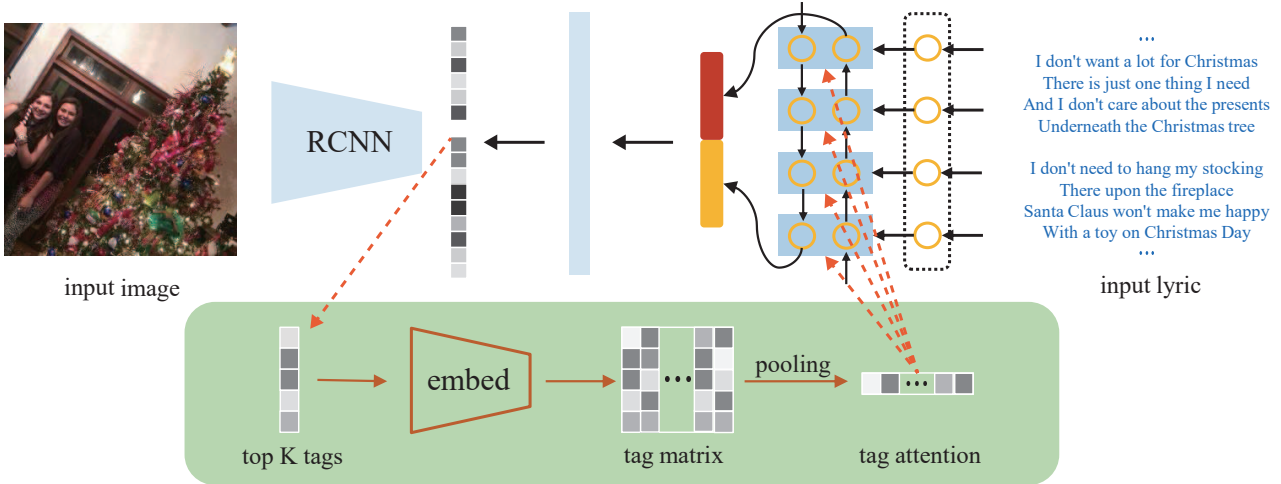}\\
\caption{The diagram of the proposed model. The image content tags are first predicted via the R-CNN, meanwhile, a bi-directional LSTM is utilized to model the corresponding lyric. Then the generated lyric representation is mapped into the space of the image tags via a MLP. To reduce the content gap between image and lyric, the top K image tags are embedded into a tag matrix and represented as tag attention for the lyric modeling by performing max/average pooling.
}\label{framework}
\end{figure*}

Considering the relevant images of a given lyric have high variance in the content, such as the examples in Fig.~\ref{dataset_ill}, it is difficult to directly correlate the image tags with each embedded lyric word, especially for the longer lyric compared with normal sentence~\cite{hermann2015teaching, tan2015lstm}.
Hence, we just take the final output of the LSTM, which remains the context of single word and provides efficient information of the whole lyric.
And experiments (Sec. 5.2) indicate its effectiveness in establishing the correlation with images.
Besides, both the forward and backward LSTM are employed to simultaneously model the history and future information of the lyric, and the final lyric representation can be denoted as
$l{\rm{ = }}{{{\rm{\mathord{\buildrel{\lower3pt\hbox{$\scriptscriptstyle\rightarrow$}} \over h} }}}_{final}} \parallel {{{\rm{\mathord{\buildrel{\lower3pt\hbox{$\scriptscriptstyle\leftarrow$}} \over h} }}}_1}$, where $\parallel$ indicates concatenation.

\subsection{Multimodal learning across image and lyric}
The image and lyric representation have been extracted via respective model, it is intuitive to embed them into a common space to establish their correlation via a~\emph{Multi-Layer Perceptron} (MLP) model, as shown in the upper part of Fig.~\ref{framework}.
The effectiveness of such model has been verified in the image-text retrieval~\cite{norouzi2013zero},
this is because the text information is specially written to describe the image content, hence there are several common features representing the same content across both modalities. However, the MLP model is not entirely suitable for our task.
As lyric is not specially generated for images, it can be found that there is few lyric words directly used to describe the corresponding image but multiple related words sharing the similar meaning with image content, which could result in a content description gap. For example, we can find such situations in the \emph{Paradise} group in Fig.~\ref{dataset_ill}. There exists few specific lyric words could be used for describing the \emph{beach}, \emph{wave}, or \emph{sky} in the images, but the words \emph{paradise}, \emph{flew}, and \emph{girl} are related to some image contents.

To address the aforementioned problem, we propose to use the tag attention to make the lyric words focus on the image content, as shown in Fig.~\ref{framework}.
We first sort the image prediction results of all the tag classes, then choose the top-K tags that are assumed as the correct prediction of corresponding image content\footnote{The top 5 tags are validated and employed in this paper.}.
To share the representation space with lyric words, the selected tags are also embedded into a vector of continuous value via Eq.~\ref{embedding}, which results in a tag matrix ${\rm{T}} \in {{\rm{R}}^{K \times 300}}$.
Then we perform the pooling operation over the matrix ${\rm{T}}$ into a tag attention vector ${\tilde {\rm{v}}}$.
Inspired by the work of Hermann~\emph{et al.}~\cite{hermann2015teaching} and Tan~\emph{et al.}~\cite{tan2015lstm}, the tag attention is designed to modify the output vector of the LSTM and make the lyric words focus on the concrete image content.
To be specific, the output vector ${\rm{h}}_t$ at time step $t$ is updated as follows,
\begin{eqnarray}\label{atten}
{{\rm{m}}_t} &=& \sigma \left( {{{\rm{W}}_{hm}}{{\rm{h}}_t} + {{\rm{W}}_{vm}}\tilde {\rm{v}}} \right)\\
{{\rm{s}}_t} &\propto& \exp \left( {{\rm{w}}_{ms}^T{{\rm{m}}_t}} \right)\\
{{\tilde {\rm{h}}}_t} &=& {{\rm{h}}_t}{{\rm{s}}_t},
\end{eqnarray}
where the weights ${\rm{W}}_{hm}$, ${\rm{W}}_{vm}$, and ${\rm{w}}_{ms}$ are considered as the attention degree of the lyric words given the image content (tags).
During modeling the lyrics, the related words are paid more attention via the attention weights, which acts like the TF-IDF in document retrieval based on key words.
However, different from the pooling operation over the entire sequence in the previous works~\cite{hermann2015teaching, tan2015lstm}, the output vector with attention, ${{\tilde {\rm{h}}}_t}$, just flow through the entire lyric words, which results in a refined lyric representation.

During training, we aims at minimizing the difference between the image and lyric pair in the tag space, which is essentially the~\emph{Mean Squared Error} (MSE),
\begin{equation}\label{loss}
{l_{mse}} = \sum\limits_{i = 1}^T {\left\| {{v_i} - {{\tilde l}_i}} \right\|_2^2},
\end{equation}
where ${v_i}$ and ${{\tilde l}_i}$ are the generated image and projected lyric representation, and $T$ is the number of training pairs.
Except the MSE loss, we also employ the cosine proximity and marginal ranking loss. The relevant experiment results are reported in the materials.
For the retrieval, both the query and retrieved items are fed into the proposed model, then the cosine similarity is computed as the relevance.

\subsection{Optimization}
The proposed model is optimized by employing stochastic gradient descent with RMSprop~\cite{tieleman2012lecture}.
The algorithm adaptively rescales the step size for updating trainable weights according to the corresponding gradient history, which achieves the best result when faced with the word frequency disparity in lyric modeling.
The parameters are empirically set as: the learning rate $l = 0.001$, the weight decay $\rho  = 0.9$, the tiny constant $\varepsilon  = {10^{ - 8}}$, and
The model is trained with mini-batches of 100 image-lyric pairs.

\section{Experiments}
\subsection{Data preprocessing and setup}
\noindent \textbf{Dataset.}
In this paper, we choose the triplets whose lyrics appear at least 5 times, which results in 8,849 triplets (586 songs). Such operation is a common preprocessing method and also better for learning the correlation across modalities.
To reduce the influence of the imbalanced number of images, we choose five triplets with top favorite counts for each song,  which are considered to have more reliable correlation.
Within these filtered triplets, 100 songs and corresponding images are randomly selected for testing, and the rest for training, which forms dataset$\dag$.
Note that, we also employ another kind of train/test partition to constitute dataset$\S$:
we randomly select one from five images of each song for testing, and the rest for training.
In the dataset$\S$, the train and test set share the same 586 songs but with different images, which is developed to exactly evaluate the models in shrinking the content gap, when faced with variable image content and lack of related lyric words.

\noindent \textbf{Preprocessing.}
For the lyric data preprocessing, we remove all the non-alphanumeric and stop words.

\noindent \textbf{Metric.} We employ the rank-based evaluation metrics. R@K is Recall@K that computes the percentage of a correct result found in the top-K retrieved items (higher is better), and Med r is the median rank of the closest correct retrieved item (lower is better).

\noindent \textbf{Baselines.}
In our experiments, we compare with the following models:
\textbf{Bag-of-Words}~\cite{bai2009polynomial}: The image features and lyric BoW representation are mapped into a common subspace.
\textbf{CONSE}~\cite{norouzi2013zero}:  The lyric representation is obtained by performing pooling operation over all words and then established correlation with image features.
\textbf{Attentive-Reader}~\cite{hermann2015teaching}:  This method is mainly developed for Question-Answering task, which performs a weighted pooling over the encoded words via LSTM with question attention. Here, the question attention is replaced with the image tag attention, and a non-linear combination is used to measure the correlation.

\noindent \textbf{Our Models.}
We evaluate two variants of our models:
\textbf{Our} baseline model: Our proposed model except the tag attention, as shown in the upper part of Fig.~\ref{framework}.
\textbf{Our-attention} model: The proposed complete model. Note that, average pooling is employed for obtaining tag attention, which could remain more image content information compared with max-manner.

\begin{table*}[t]
\begin{center}
\renewcommand\arraystretch{1}
\begin{tabular}{C{3.1cm}|C{0.6cm}C{0.6cm}C{0.9cm}C{0.9cm}|C{0.6cm}C{0.6cm}C{0.9cm}C{0.9cm}|C{0.6cm}C{0.6cm}C{0.9cm}|C{0.9cm}}
  \hline
  Image tags &  \multicolumn{4}{c|}{obj-tags} & \multicolumn{4}{c|}{attr-tags} & \multicolumn{4}{c}{obj-attr-tags} \\ \hline \hline
  Dataset$\dag$    & R@1  & R@5  & R@10     & Med r       & R@1    & R@5  & R@10 & Med r                        & R@1    & R@5  & R@10  & Med r        \\ \hline
BoW~\cite{bai2009polynomial}     &      1.4   &	7.6& 13.4  & 	45.06                                            &      1.6   &	7.0	&  13.0  & 	46.11                                          &     1.4    &	7.2	&  13.2  & 	45.01     \\ 
CONSE~\cite{norouzi2013zero}    &     1.8   &	 8.0 	&  13.0  & 	46.11                                            &      1.8   &	6.6	&  12.2  & 	47.38                                          &     2.0   &	6.8	 &  14.8  & 	46.13                            \\ 
Attentive-Reader~\cite{hermann2015teaching}   &      1.8   &	 7.4	&  14.0  & 	44.45                        &     1.8   &	 7.6	&  13.0  & 	46.97                                          &      1.8  &	 7.2	&  15.0  & 	43.78         \\ \hline
Our &    \textbf{2.2}	  & 	7.2	 & 14.2   &	43.13        &   1.6	 &	7.8 &  14.0   &	 46.32                                         &    2.2	    & 	7.6 & 14.8  &	43.30  \\
Our-attention & 2.0  	& \textbf{9.2} &	\textbf{17.6}     &	\textbf{41.36}        & \textbf{2.2}  	& \textbf{8.4} &	   \textbf{14.4}   &	\textbf{45.21}            & \textbf{2.6}   & \textbf{9.4} &	\textbf{16.8} & \textbf{41.50} \\ \hline \hline
  Dataset$\S$     & R@10  & R@50  & R@100     & Med r       & R@10    & R@50  & R@100 & Med r                        & R@10    & R@50  & R@100  & Med r        \\ \hline
BoW~\cite{bai2009polynomial}     &      4.27   &	13.48	&  24.23  & 	251.77                                            &      3.07   &	 12.12	&  23.55  & 	260.17                                     &   4.27   &	14.33	&  23.72  & 	250.63                           \\ 
CONSE~\cite{norouzi2013zero}    &     3.41   &	 14.33	&  25.27  & 	248.88                                            &      3.24   &	 12.97	&  23.38  & 	262.17                                          &     3.58   &	14.51	&  25.09  & 	245.32                            \\ 
Attentive-Reader~\cite{hermann2015teaching}   &      4.10   &	14.51	& 24.57  & 	254.53                                            &     3.75   &	 13.14	&  23.04  & 267.12                     &      3.92   &	14.51	&  25.43  & 	243.53         \\ \hline
Our &    \textbf{4.61}	  & 	15.02	 &  26.28   &	246.51        &   4.27	 &	13.65 & 23.89   &	 254.17           &    4.94	    & 	15.36	 &  28.33  &	240.00 \\
Our-attention & 4.47  	& \textbf{15.36} &	\textbf{26.86}     &	\textbf{245.61}        &  \textbf{4.61}  	& \textbf{13.99} &	\textbf{25.76}      &	\textbf{249.42}            & \textbf{5.63}   & \textbf{17.58} &	 \textbf{29.35} &	 \textbf{233.82} \\ \hline
\end{tabular}
\end{center}
\caption{\label{table1}Image2song retrieval experiment result in R@K and Med r on dataset$\dag$ and dataset$\S$. Three kinds of image representation are considered, \emph{e.g.}, object (obj), attribute (attr), and both them (obj-attr).}
\end{table*}

\begin{figure*}[t]
\centering
\includegraphics[width=17cm]{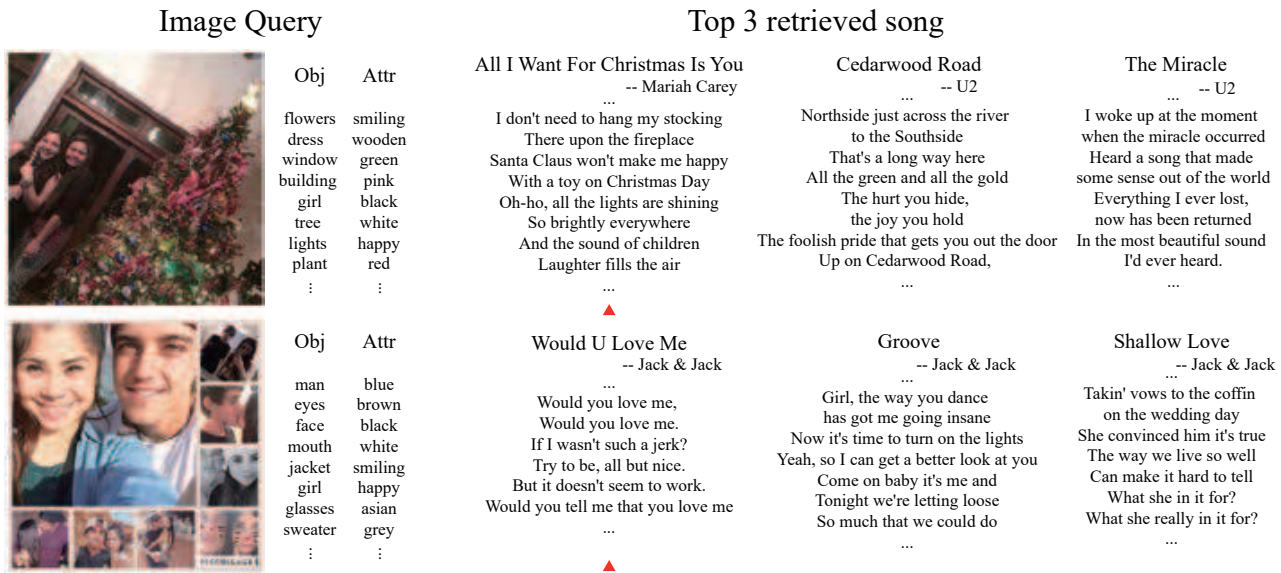}\\
\caption{Image2song retrieval examples generated by our model. The generated object and attribute tags are shown next to each image query, and the songs with red triangle are the ground truth in the dataset. }\label{resultImage}
\end{figure*}

\subsection{Image2song retrieval}
This experiment aims to evaluate the song retrieval performance by given image query.
There are two kinds of tags used for representing images, \emph{i.e.}, object and attribute. It is expected to explore which kind of them influences the image description most and provides more valuable information to establish the correlation across the two modalities. In view of this, we group the image tags into three categories, \emph{i.e.}, object, attribute, and both of them.
Table.~\ref{table1} shows the comparison results on both datasets, where both the model variants and other methods are considered. And we also provide example retrieval results in Fig.~\ref{resultImage}.

For each tag category, there are three points we should pay attention to.
First, the bi-directional LSTM model provides better lyric representations than direct pooling (CONSE~\cite{norouzi2013zero}) and BoW~\cite{bai2009polynomial}, as the LSTM takes consideration of both word semantic and context information.
Second, the proposed complete model shows the best result in most conditions.
When we employ the tag attention for lyric modeling, more related words in the lyrics will be emphasized, which shrinks the content gap and improves the relevant performance, especially on the dataset$\S$.
Although Attentive-Reader~\cite{hermann2015teaching} also employ similar manner, it takes the attentive pooling result as the lyric representation.
The direct pooling operation in Attentive-Reader will make it suffer from the difficulty in establishing correlation,
this is because the variable image contents change the attention weight.
While our model takes the final output of LSTM, where the attention weight is not immediately utilized but conveyed by the updated output vector ${\rm{\tilde h}}_t$.
Third, although our methods show the best performance on both datasets, overall performance is not excellent as expected.
But this is a common situation in the semantic-based music retrieval~\cite{schedl2014music}.
This is because the song retrieval based on textual data has to estimate the semantic labels from lyric, which is charactered as a low specificity and long-term granularity~\cite{schedl2014music}.
Even so, our proposed models still enjoy a relatively big improvement, and the retrieved examples show the effectiveness of the models in Fig.~\ref{resultImage}.

Across the three groups of image tags, we could find that the attribute tags almost always perform worse than object ones. We consider there are mainly three reasons to explain this phenomenon.
First, the object words are usually employed for image description, which is more powerful in identifying the image content compared with attribute ones.
Second, as the Shuttersong is actually a kind of sharing application, most of the updated images are self-photography, which makes it difficult to correlate images with songs, especially for the attribute tags.
Third, most of the images share similar attribute tags with high prediction scores (\emph{e.g.}, $black$, $blue$, $white$). This is actually a long-tailed distribution\footnote{A detailed illustration can be found in the supplementary material.} and therefore it becomes difficult to establish the correlation between image and specific lyric.
However, the group with both tags nearly shows the best performance across different models, which results from more detailed description for images.

\begin{figure}[t]
\centering
\includegraphics[width=8cm]{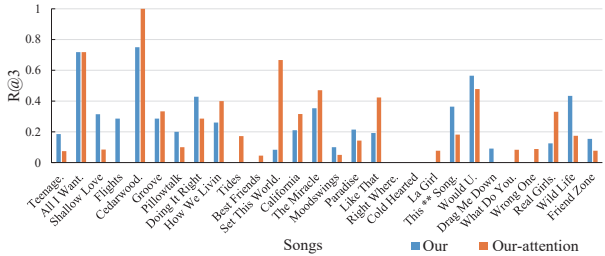}\\
\caption{Detailed comparison results among song examples in R@3. The complete model with attention improves the average performance, especially for the songs with \textbf{zero} score.}\label{attention}
\end{figure}

\begin{figure}[t]
\centering
\includegraphics[width=8cm]{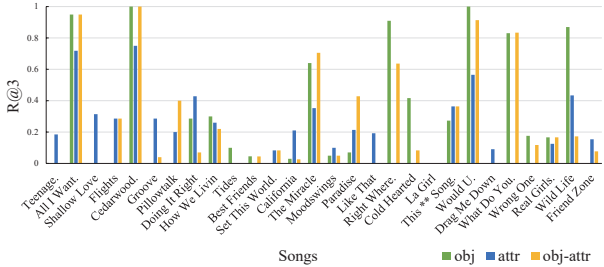}\\
\caption{Detailed comparison results among the three groups of tags. All the experiments are conducted with the proposed baseline model and evaluated by R@3.}\label{tags}
\end{figure}

Apart from the influence of image tag property, the lyric also impacts the performance significantly.
We show the specific results of 28 songs with more 50 times occurrence in Fig.~\ref{attention} and Fig.~\ref{tags}.
As shown, some song lyrics are of remarkable performance, while some fail to establish the correlation with relevant images.
The potential reason is that parts of the lyrics are weak in providing obvious or even related words.
For example, the lyric words of  \emph{Best Friends} are about \emph{forget}, \emph{tonight}, \emph{remember}, etc, which are not specifically related to the image content, as shown in Fig.~\ref{dataset_ill}.
Even so, our proposed tag attention mechanism can still reduce the content gap and improve the performance, as show in Fig.~\ref{attention}.
While for the song \emph{Paradise} and \emph{All I Want For Christmas Is You}, the lyric words and image content are closely related, hence this case achieves better results.

\begin{table}[t]
\begin{center}
\begin{tabular}{c|c|c|c|c|c}
\hline
Models    & R@1  & R@5  & R@10  & R@20     &   Med r      \\ \hline
Schwarz~\cite{schwarz2016autoIllustration}   & 0.10 & 0.27 & 0.45 & 0.71 &   16          \\
Our   & 0.19 & 0.34 & 0.52 & 0.74 &    15          \\ \hline
\end{tabular}
\end{center}
\caption{\label{table2} The image retrieval results given lyric query. Here, the image tags in the method of Schwarz~\emph{et al.}~\cite{schwarz2016autoIllustration} are generated by the R-CNN approach~\cite{ren2015faster} for a fair comparison.}
\end{table}

\subsection{Song2image retrieval}
In this experiment, we aims to retrieve relevant images for a given song (lyric) query, which is similar to the music visualization task.
And we employ the proposed baseline model, which could perform more efficiently without the attention of each image.
Table.~\ref{table2} shows the comparison results on dataset$\dag$.
First, our proposed model outperforms the method of Schwarz~\emph{et al.}~\cite{schwarz2016autoIllustration}. In our model, the image content information are employed to supervise the lyric modeling during training,
while Schwarz~\emph{et al.}~\cite{schwarz2016autoIllustration} directly use the pooling results of lyric word vector for similarity retrieval without the inner-model interaction like ours.
Second, although our model get better performance, it still suffers from the lack of related words in some songs, just like the image2song task.

\begin{figure}[t]
\centering
\includegraphics[width=8cm]{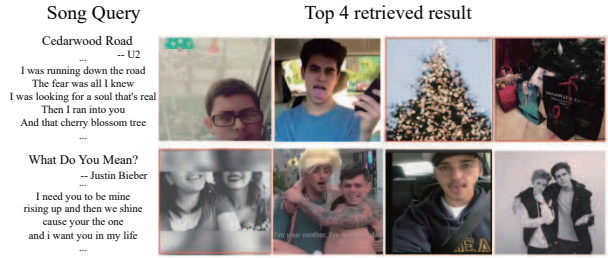}\\
\caption{Example retrieval results by our model. The images in red bounding box are the corresponding ones in the Dataset.}\label{resultLyric}
\end{figure}

In addition, we also show some examples of the song query, and the top 4 retrieved results are illustrated in Fig.~\ref{resultLyric}.
Although some retrieved results are not correct, they share similar content conveyed by the lyric.
For example, in terms of the song \emph{Cedarwood Road} about tree and Christmas, the top two retrieved images indeed have tree-relevant content and the left two are about Christmas.

\section{Discussion}
In this paper, we introduce a novel problem that retrieves semantic-related songs based on given images, which is named as the image2song task.
We collect a dataset that consists of pairwise images and songs to study this problem.
We propose a semantic-based song retrieval framework, which employs the lyric as the textual data source for estimating the semantic label of songs,
then a deep neural network based multimodal framework is proposed to learn the correlation, where the lyric modeling is proposed to focus on the main image content to reduce the content gap between them.
The experiment results show that our model can recommend suitable songs for a given image.
In addition, the proposed approach can also retrieve relevant images according to a song query with a better performance than other methods.

There still remains a direction that should be explored in the future.
Song is about several minutes long, which is too long for just showing an image.
A possible way could be expected is to correlate the image with only parts of the corresponding song, which is more natural for expression.
Furthermore, we also hope to perform other kinds of cross-modal retrieval task, which essentially attempts to establish the correlation among different senses of human.

\section{Acknowledgement}
We thank Yaxing Sun for crawling the raw multimodal data from the Shuttersong application.
We also thank Chengtza Wang for editing the video demo for presentation.

\bibliographystyle{ieee}
\bibliography{image2song}

\begin{thebibliography}{10}\itemsep=-1pt

\bibitem{bai2009polynomial}
B.~Bai, J.~Weston, D.~Grangier, R.~Collobert, K.~Sadamasa, Y.~Qi, C.~Cortes,
  and M.~Mohri.
\newblock Polynomial semantic indexing.
\newblock In {\em Advances in Neural Information Processing Systems}, pages
  64--72, 2009.

\bibitem{cai2007automated}
R.~Cai, L.~Zhang, F.~Jing, W.~Lai, and W.-Y. Ma.
\newblock Automated music video generation using web image resource.
\newblock In {\em 2007 IEEE International Conference on Acoustics, Speech and
  Signal Processing-ICASSP'07}, volume~2, pages II--737. IEEE, 2007.

\bibitem{celma2006search}
{\`O}.~Celma, P.~Cano, and P.~Herrera.
\newblock Search sounds an audio crawler focused on weblogs.
\newblock In {\em 7th International Conference on Music Information Retrieval
  (ISMIR)}, 2006.

\bibitem{chen2006content}
R.~Chen, Z.~Xu, Z.~Zhang, and F.~Luo.
\newblock Content based music emotion analysis and recognition.
\newblock In {\em Proc. of 2006 International Workshop on Computer Music and
  Audio Technology}, volume 68275, 2006.

\bibitem{chu2016song}
H.~Chu, R.~Urtasun, and S.~Fidler.
\newblock Song from pi: A musically plausible network for pop music generation.
\newblock {\em arXiv preprint arXiv:1611.03477}, 2016.

\bibitem{dong2016word2visualvec}
J.~Dong, X.~Li, and C.~G. Snoek.
\newblock Word2visualvec: Cross-media retrieval by visual feature prediction.
\newblock {\em arXiv preprint arXiv:1604.06838}, 2016.

\bibitem{everingham2010pascal}
M.~Everingham, L.~Van~Gool, C.~K. Williams, J.~Winn, and A.~Zisserman.
\newblock The pascal visual object classes (voc) challenge.
\newblock {\em International journal of computer vision}, 88(2):303--338, 2010.

\bibitem{graves2012neural}
A.~Graves.
\newblock Neural networks.
\newblock In {\em Supervised Sequence Labelling with Recurrent Neural
  Networks}, pages 15--35. Springer, 2012.

\bibitem{graves2013speech}
A.~Graves, A.-r. Mohamed, and G.~Hinton.
\newblock Speech recognition with deep recurrent neural networks.
\newblock In {\em 2013 IEEE international conference on acoustics, speech and
  signal processing}, pages 6645--6649. IEEE, 2013.

\bibitem{hermann2015teaching}
K.~M. Hermann, T.~Kocisky, E.~Grefenstette, L.~Espeholt, W.~Kay, M.~Suleyman,
  and P.~Blunsom.
\newblock Teaching machines to read and comprehend.
\newblock In {\em Advances in Neural Information Processing Systems}, pages
  1693--1701, 2015.

\bibitem{hochreiter1997long}
S.~Hochreiter and J.~Schmidhuber.
\newblock Long short-term memory.
\newblock {\em Neural computation}, 9(8):1735--1780, 1997.

\bibitem{hu2016multimodal}
D.~Hu, X.~Lu, and X.~Li.
\newblock Multimodal learning via exploring deep semantic similarity.
\newblock In {\em Proceedings of the 2016 ACM on Multimedia Conference}, pages
  342--346. ACM, 2016.

\bibitem{hu2009lyric}
X.~Hu, J.~S. Downie, and A.~F. Ehmann.
\newblock Lyric text mining in music mood classification.
\newblock {\em American music}, 183(5,049):2--209, 2009.

\bibitem{johnson2015image}
J.~Johnson, R.~Krishna, M.~Stark, L.-J. Li, D.~A. Shamma, M.~S. Bernstein, and
  L.~Fei-Fei.
\newblock Image retrieval using scene graphs.
\newblock In {\em 2015 IEEE Conference on Computer Vision and Pattern
  Recognition (CVPR)}, pages 3668--3678. IEEE, 2015.

\bibitem{karpathy2015deep}
A.~Karpathy and L.~Fei-Fei.
\newblock Deep visual-semantic alignments for generating image descriptions.
\newblock In {\em Proceedings of the IEEE Conference on Computer Vision and
  Pattern Recognition}, pages 3128--3137, 2015.

\bibitem{kim2010music}
Y.~E. Kim, E.~M. Schmidt, R.~Migneco, B.~G. Morton, P.~Richardson, J.~Scott,
  J.~A. Speck, and D.~Turnbull.
\newblock Music emotion recognition: A state of the art review.
\newblock In {\em Proc. ISMIR}, pages 255--266. Citeseer, 2010.

\bibitem{kiros2014multimodal}
R.~Kiros, R.~Salakhutdinov, and R.~S. Zemel.
\newblock Multimodal neural language models.
\newblock In {\em ICML}, volume~14, pages 595--603, 2014.

\bibitem{kiros2014unifying}
R.~Kiros, R.~Salakhutdinov, and R.~S. Zemel.
\newblock Unifying visual-semantic embeddings with multimodal neural language
  models.
\newblock {\em arXiv preprint arXiv:1411.2539}, 2014.

\bibitem{knees2007music}
P.~Knees, T.~Pohle, M.~Schedl, and G.~Widmer.
\newblock A music search engine built upon audio-based and web-based similarity
  measures.
\newblock In {\em Proceedings of the 30th annual international ACM SIGIR
  conference on Research and development in information retrieval}, pages
  447--454. ACM, 2007.

\bibitem{kulkarni2013babytalk}
G.~Kulkarni, V.~Premraj, V.~Ordonez, S.~Dhar, S.~Li, Y.~Choi, A.~C. Berg, and
  T.~L. Berg.
\newblock Babytalk: Understanding and generating simple image descriptions.
\newblock {\em IEEE Transactions on Pattern Analysis and Machine Intelligence},
  35(12):2891--2903, 2013.

\bibitem{li2007and}
L.-J. Li and L.~Fei-Fei.
\newblock What, where and who? classifying events by scene and object
  recognition.
\newblock In {\em 2007 IEEE 11th International Conference on Computer Vision},
  pages 1--8. IEEE, 2007.

\bibitem{li2008visual}
X.~Li, D.~Tao, S.~J. Maybank, and Y.~Yuan.
\newblock Visual music and musical vision.
\newblock {\em Neurocomputing}, 71(10):2023--2028, 2008.

\bibitem{Liem2012WhenMM}
C.~C.~S. Liem, M.~Larson, and A.~Hanjalic.
\newblock When music makes a scene.
\newblock {\em IJMIR}, 2:15--30, 2012.

\bibitem{lin2014microsoft}
T.-Y. Lin, M.~Maire, S.~Belongie, J.~Hays, P.~Perona, D.~Ramanan,
  P.~Doll{\'a}r, and C.~L. Zitnick.
\newblock Microsoft coco: Common objects in context.
\newblock In {\em European Conference on Computer Vision}, pages 740--755.
  Springer, 2014.

\bibitem{Long2015FullyCN}
J.~Long, E.~Shelhamer, and T.~Darrell.
\newblock Fully convolutional networks for semantic segmentation.
\newblock {\em CoRR}, abs/1411.4038, 2015.

\bibitem{mao2014deep}
J.~Mao, W.~Xu, Y.~Yang, J.~Wang, Z.~Huang, and A.~Yuille.
\newblock Deep captioning with multimodal recurrent neural networks (m-rnn).
\newblock {\em arXiv preprint arXiv:1412.6632}, 2014.

\bibitem{mikolov2013distributed}
T.~Mikolov, I.~Sutskever, K.~Chen, G.~S. Corrado, and J.~Dean.
\newblock Distributed representations of words and phrases and their
  compositionality.
\newblock In {\em Advances in neural information processing systems}, pages
  3111--3119, 2013.

\bibitem{miyazaki2010dynamicicon}
R.~Miyazaki and K.~Matsuda.
\newblock Dynamicicon: A visualizing technique for musical pieces in moving
  icons based on acoustic features.
\newblock {\em Journal of Information Processing}, 51(5):1283--1293, 2010.

\bibitem{norouzi2013zero}
M.~Norouzi, T.~Mikolov, S.~Bengio, Y.~Singer, J.~Shlens, A.~Frome, G.~S.
  Corrado, and J.~Dean.
\newblock Zero-shot learning by convex combination of semantic embeddings.
\newblock {\em arXiv preprint arXiv:1312.5650}, 2013.

\bibitem{ren2015faster}
S.~Ren, K.~He, R.~Girshick, and J.~Sun.
\newblock Faster r-cnn: Towards real-time object detection with region proposal
  networks.
\newblock In {\em Advances in neural information processing systems}, pages
  91--99, 2015.

\bibitem{russakovsky2015imagenet}
O.~Russakovsky, J.~Deng, H.~Su, J.~Krause, S.~Satheesh, S.~Ma, Z.~Huang,
  A.~Karpathy, A.~Khosla, M.~Bernstein, et~al.
\newblock Imagenet large scale visual recognition challenge.
\newblock {\em International Journal of Computer Vision}, 115(3):211--252,
  2015.

\bibitem{schedl2014music}
M.~Schedl, E.~G{\'o}mez, J.~Urbano, et~al.
\newblock Music information retrieval: Recent developments and applications.
\newblock {\em Foundations and Trends{\textregistered} in Information
  Retrieval}, 8(2-3):127--261, 2014.

\bibitem{schwarz2016autoIllustration}
K.~Schwarz, T.~L. Berg, and H.~P.~A. Lensch.
\newblock Auto-illustrating poems and songs with style.
\newblock In {\em Asian Conference on Computer Vision (ACCV)}, 2016.

\bibitem{simonyan2014very}
K.~Simonyan and A.~Zisserman.
\newblock Very deep convolutional networks for large-scale image recognition.
\newblock {\em arXiv preprint arXiv:1409.1556}, 2014.

\bibitem{sivic2005discovering}
J.~Sivic, B.~C. Russell, A.~A. Efros, A.~Zisserman, and W.~T. Freeman.
\newblock Discovering object categories in image collections.
\newblock 2005.

\bibitem{sivic2003video}
J.~Sivic and A.~Zisserman.
\newblock Video google: A text retrieval approach to object matching in videos.
\newblock In {\em Computer Vision, 2003. Proceedings. Ninth IEEE International
  Conference on}, pages 1470--1477. IEEE, 2003.

\bibitem{sohn2014improved}
K.~Sohn, W.~Shang, and H.~Lee.
\newblock Improved multimodal deep learning with variation of information.
\newblock In {\em Advances in Neural Information Processing Systems}, pages
  2141--2149, 2014.

\bibitem{sordo2009querybag}
M.~Sordo, {\`O}.~Celma, and C.~Laurier.
\newblock Querybag: Using different sources for querying large music
  collections.
\newblock In {\em Proceedings of the 10th International Society for Music
  Information Retrieval Conference (ISMIR)}, 2009.

\bibitem{srivastava2012multimodal}
N.~Srivastava and R.~R. Salakhutdinov.
\newblock Multimodal learning with deep boltzmann machines.
\newblock In {\em Advances in neural information processing systems}, pages
  2222--2230, 2012.

\bibitem{stein1993merging}
B.~E. Stein and M.~A. Meredith.
\newblock {\em The merging of the senses.}
\newblock The MIT Press, 1993.

\bibitem{sutskever2014sequence}
I.~Sutskever, O.~Vinyals, and Q.~V. Le.
\newblock Sequence to sequence learning with neural networks.
\newblock In {\em Advances in neural information processing systems}, pages
  3104--3112, 2014.

\bibitem{tan2015lstm}
M.~Tan, B.~Xiang, and B.~Zhou.
\newblock Lstm-based deep learning models for non-factoid answer selection.
\newblock {\em arXiv preprint arXiv:1511.04108}, 2015.

\bibitem{tieleman2012lecture}
T.~Tieleman and G.~Hinton.
\newblock Lecture 6.5-rmsprop: Divide the gradient by a running average of its
  recent magnitude.
\newblock {\em COURSERA: Neural Networks for Machine Learning}, 4(2), 2012.

\bibitem{van2010automatic}
M.~Van~Zaanen and P.~Kanters.
\newblock Automatic mood classification using tf* idf based on lyrics.
\newblock In {\em ISMIR}, pages 75--80, 2010.

\bibitem{Wang2007RetrievingWI}
Z.-K. Wang, R.~Cai, L.~Zhang, Y.~Zheng, and J.-M. Li.
\newblock Retrieving web images to enrich music representation.
\newblock In {\em ICME}, 2007.

\bibitem{wu2015value}
Q.~Wu, C.~Shen, L.~Liu, A.~Dick, and A.~v.~d. Hengel.
\newblock What value do explicit high level concepts have in vision to language
  problems?
\newblock {\em arXiv preprint arXiv:1506.01144}, 2015.

\bibitem{xia2008sentiment}
Y.~Xia, L.~Wang, and K.-F. Wong.
\newblock Sentiment vector space model for lyric-based song sentiment
  classification.
\newblock {\em International Journal of Computer Processing Of Languages},
  21(04):309--330, 2008.

\bibitem{xu2015show}
K.~Xu, J.~Ba, R.~Kiros, K.~Cho, A.~Courville, R.~Salakhutdinov, R.~S. Zemel,
  and Y.~Bengio.
\newblock Show, attend and tell: Neural image caption generation with visual
  attention.
\newblock {\em arXiv preprint arXiv:1502.03044}, 2(3):5, 2015.

\bibitem{Xu2008AutomaticGO}
S.~Xu, T.~Jin, and F.~C.-M. Lau.
\newblock Automatic generation of music slide show using personal photos.
\newblock In {\em ISM}, 2008.

\bibitem{yoshii2008music}
K.~Yoshii and M.~Goto.
\newblock Music thumbnailer: Visualizing musical pieces in thumbnail images
  based on acoustic features.
\newblock In {\em ISMIR}, pages 211--216, 2008.

\end{thebibliography}

\clearpage

\noindent{\Large{\textbf{Supplementary Material}}}
\section{The Shuttersong Dataset}
\subsection{Favorite Count}
\begin{figure}[h]
\centering
\includegraphics[width=5cm]{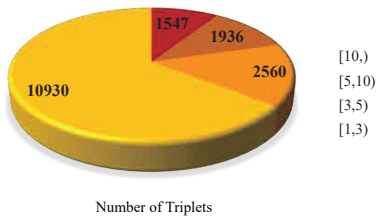}\\
\caption{The statistics of triplet number in favorite counts. There are 1,547 triplets owning at least 10 favorite counts, which could be considered as the image-song pair with high quality.}\label{favorite}
\end{figure}

Apart from the song clip, image, and mood, we also collect the favorite count for each image-song pair from the Shuttersong application.
The favorite counts vary from 1 to 8,964, which could be used to estimate the quality of image-song pairs as a reference.
The specific statistics can be found in Fig.~\ref{favorite}.
There are 6,043 (image, music clip, lyric) triplets owning at least 3 favorite counts, which are considered to jointly show better expressions compared with the others.

\subsection{Lyric Refinement}
\begin{figure*}[t]
\centering
\includegraphics[width=16cm]{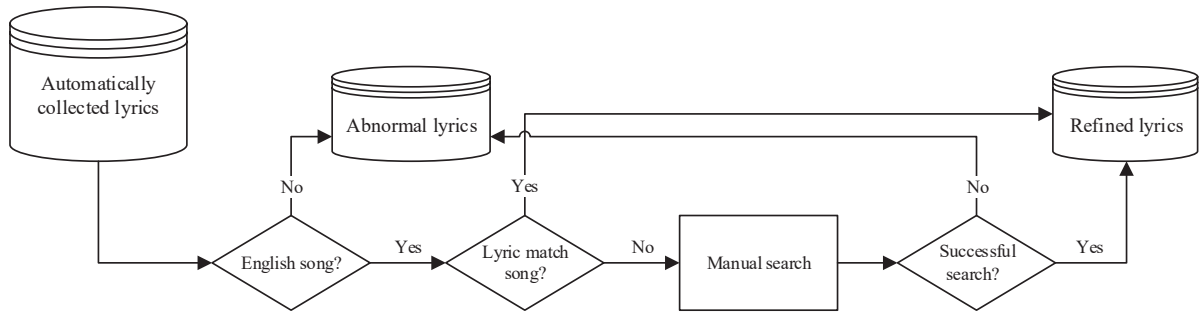}\\
\caption{The flow chart of manual lyric refinement. The automatically collected lyrics are divided into two parts, one is the abnormal ones that contains non-English song and undetected lyric, while the other is the
refined lyrics used to constitute the final Shuttersong dataset.}\label{lyric}
\end{figure*}

As there are some abnormal lyrics existing in the automatically searched set, it is necessary to verify each of them. Hence, we ask twenty participants to refine the lyrics, and the corresponding flow char of the refinement is shown in Fig.~\ref{lyric}. First, the participants judge whether the song is in English or not.  Then they select the mismatch ones and conduct manual searching for the filtered English songs. The websites used for searching in this paper are \emph{www.musixmatch.com} and \emph{search.azlyrics.com}. Finally, both the correct matching and successfully updated ones constitute the refined lyric set. And the rest lyrics are the abnormal ones, \emph{e.g.} non-English songs, unfound lyrics.

\begin{table*}[t]
\begin{center}
\renewcommand\arraystretch{1.2}
\begin{tabular}{C{3.1cm}|C{0.7cm}|C{0.7cm}|C{0.7cm}|C{0.9cm}|C{0.7cm}|C{0.7cm}|C{0.7cm}|C{0.9cm}|C{0.7cm}|C{0.7cm}|C{0.7cm}|C{0.9cm}}
  \hline
  Image tags &  \multicolumn{4}{c|}{obj-tags} & \multicolumn{4}{c|}{attr-tags} & \multicolumn{4}{c}{obj-attr-tags} \\ \hline
  Models    & R@1  & R@5  & R@10     & Med r       & R@1    & R@5  & R@10 & Med r                        & R@1    & R@5  & R@10  & Med r        \\ \hline
BoW~\cite{bai2009polynomial}   &      10.71   &	31.21	& 52.62  & 	9.34                                            &      9.32   &	30.03	&  51.34  & 	10.06                                          &    9.42     &	34.51	&  55.73  & 	9.15                           \\ 
CONSE~\cite{norouzi2013zero}     &     10.44   &	 30.93	&  52.42  & 	9.50                                            &      9.13   &	29.61	&  51.19  & 	10.20                                          &     9.39   &	34.24	&  55.19  & 	9.35                            \\ 
Attentive-Reader~\cite{hermann2015teaching}   &      11.45   &	32.81	& 52.02  & 	9.47                                            &     9.13   &	 30.26	&  51.47  & 	9.91                                          &      12.95   &	37.16	&  61.79  & 	8.62         \\ \hline
Our &    11.34	  & 	32.52	 & 51.44   &	9.61        &   8.92	 &	29.82 & 51.18   &	 9.81           &    10.95	    & 	36.31	 & 57.51  &	8.87 \\
Our-mood & 12.13 & 34.52 & 54.60  &	8.83          & \textbf{9.70}   & 31.31 & \textbf{52.84}       &	9.13              & 12.13   & 37.46 & 61.85     &	 8.23        \\ 
Our-attention & \textbf{12.71}  	& \textbf{35.14} &	\textbf{57.37}     &	\textbf{8.37}        & 9.26  	& \textbf{33.64} &	52.26      &	\textbf{8.97}            & \textbf{13.10}   & \textbf{38.38} &	\textbf{62.50} &	 \textbf{7.82} \\ \hline
\end{tabular}
\end{center}
\caption{\label{table1}Image2song retrieval experiment result in R@K and Med r. Three kinds of image representation are considered, \emph{e.g.}, object (obj), attribute (attr), and both them (obj-attr).}
\end{table*}

\begin{figure}[h]
\centering
\includegraphics[width=8cm]{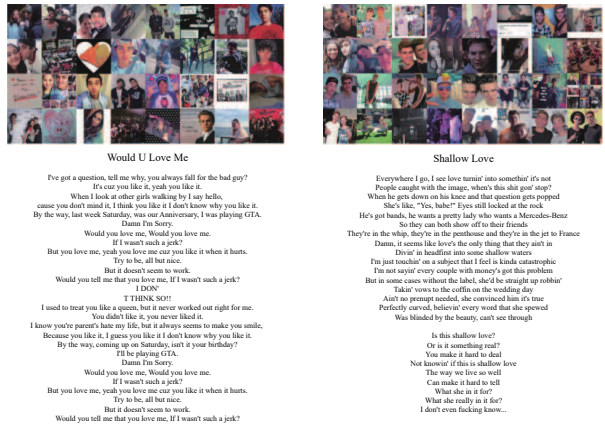}\\
\caption{Examples of songs with high frequency appearance in the Shuttersong dataset. Multiple corresponding images are also shown for each of them.}\label{images}
\end{figure}

\section{Additional Experiments}
We have shown the specific comparison results of the 28 songs with more 50 times occurrence in the paper, The following subsections show more results of our models with these songs, as well as other compared models.

\subsection{More Retrieval Results}
Apart from the lyric words and image features, we also take consideration of the mood information, which is combined with the encoded lyric representation, but only 18.6\% is available.
As shown in Table~\ref{table1}, the extra mood information indeed strengthens the correlation between image and lyric, which even outperforms the attention model in some cases. This is because the mood tag directly points out the core information of the shared image-song pair and therefore makes the pair become closer.

\subsection{Pooling Operation}
The tag attention is obtained by performing the pooling operation over the tag matrix, which plays an important role in establishing the correlation between image and lyric.
In view of this, the average and max pooling strategy are compared to evaluate their performances in remaining effective image content.
Table.~\ref{pooling} shows the comparison results.
It is clear that using average pooling is much better than max pooling. The potential reason is that the average pooling could extract more tag semantic values from the tag matrix, so that more tag values provide a more complete description for images.

\begin{table}[h]
\begin{center}
\begin{tabular}{c|c|c|c|c|c}
\hline
Pooling   & R@1  & R@3  & R@5  & R@10           &   Med r      \\ \hline
Average   &  13.10     & 28.30 & 38.38     & 62.50 	&  7.82          \\
Max         & 12.08      & 26.54 & 35.40     & 59.74  &   8.37          \\\hline
\end{tabular}
\end{center}
\caption{\label{pooling}The performance of the proposed model with different pooling strategies over the tag matrix.}
\end{table}

\begin{table*}[t]
\begin{center}
\begin{tabular}{c|c|c|c|c|c|c|c|c|c|c}
\hline
Attributes                      &  white  & black  & blue  &  brown     &   red  & green  & pink  & blonde & smiling   &   $ \cdots$         \\ \hline
Average Probabilities  &   0.30   &	0.25  &  0.20   &	0.19 	&  0.14  &0.12  &0.09  &0.09  & 0.08  &   $ \cdots$         \\ \hline
\end{tabular}
\end{center}
\caption{\label{table2} The top 9 detected attributes with corresponding prediction probabilities.}
\end{table*}

\subsection{Loss Comparison}
In addition to the~\emph{Mean Squared Error} (MSE) loss function employed in the paper,  \emph{Cosine Proximity Loss} (CPL) and \emph{Marginal Ranking Loss} (MRL)  are also considered.
CPL is based on the cosine distance, which is commonly used in vector space model and written as follow,
\begin{equation}\label{cpl}
{l_{cpl}} =  - \sum\limits_{i = 1}^T {\cos \left( {{v_i},~{{\tilde l}_i}} \right)}.
\end{equation}
As for MRL, it takes consideration of both positive and negative samples with respect to the images query and is more prevalent in retrieval tasks.
It belongs to the hinge loss and is written as,
\begin{equation}\label{mrl}
{l_{mrl}} = \sum\limits_{i = 1}^T {\max \left\{ {0,1 + \cos \left( {{v_i},~\tilde l_i^ - } \right) - \cos \left( {{v_i},~\tilde l_i^ + } \right)} \right\}},
\end{equation}
where ${\tilde l_i^ +}$ is the ground truth lyric for current image representation $v_i$, and ${\tilde l_i^ -}$ is a negative one that is randomly selected from the entire lyric database.

Table.~\ref{loss} shows the comparison results among the three introduced loss functions. It is obvious that MSE performs the best in both Recall@K and Med r metric, while MRL has the worst performance. We consider that the main reason comes from the diversity of images, \emph{e.g.} the examples in Fig.~\ref{images}. The images related to the same lyrics have high variance in the appearance, which makes these two modalities lack the content correspondence to each other. Hence, it becomes more challenging to deal with the positive and negative samples simultaneously. Such conditions can be also found in the image-to-text retrieval task~\cite{dong2016word2visualvec}.

\begin{table}[h]
\begin{center}
\begin{tabular}{c|c|c|c|c|c}
\hline
Loss    & R@1  & R@3  & R@5  & R@10     &   Med r        \\ \hline
MRL   & 9.90    & 22.70  & 36.04     & 57.84    &      8.94        \\
CPL   & 11.29   & 26.25  & 37.07     & 60.92    &      8.29        \\
MSE  &  13.10  & 28.30  & 38.38     & 62.50    &      7.82        \\\hline
\end{tabular}
\end{center}
\caption{\label{loss} The retrieval performance of our model with distinct loss functions.}
\end{table}

\begin{figure}[t]
\centering
\includegraphics[width=8cm]{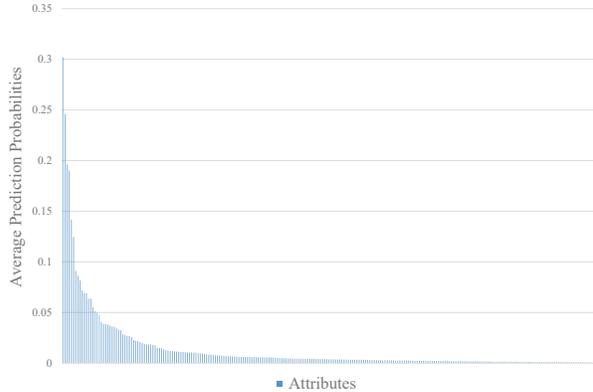}\\
\caption{The average attribute prediction results over all the images in dataset$\dag$. The results are sorted in the descend order. }\label{longtail}
\end{figure}

\subsection{Attribute Property}
In our paper, the attribute tags perform worse than the object ones, one of the potential reasons is due to the imbalanced attributes.
We perform a statistical analysis with the attribute prediction probabilities, where all the images whose corresponding lyrics appear at least 5 times are considered.
There are 249 attribute types employed in this paper, and Fig.~\ref{longtail} shows the average prediction results.
It is clear to find that only a few types have high value, while most remain the low probabilities, which is actually a kind of long-tailed distribution.
The imbalanced results could make it difficult to distinguish the images that belong to different songs.
More importantly, the top 9 attributes are almost color-related, as shown in Table.~\ref{table2}.
These attributes commonly appear in colorful images, and therefore become weaker in describing the specific image appearance compared with other ones, \emph{e.g.}~\emph{happy},~\emph{messy}.
Hence, only employing attribute tags may suffer from the aforementioned problems and result in the unreliable correlation.

\end{document}